\documentclass[twocolumn,10pt]{asme2e}

\usepackage{graphicx}
\usepackage{cite}
\usepackage[bottom]{footmisc}
\usepackage{psfrag}
\usepackage{floatflt}
\usepackage{enumerate}
\usepackage{amsfonts}
\usepackage{color}
\usepackage{amsmath}

\usepackage{array,contour}
\newcommand{\gbf}[1] {\contour[5]{black}{${#1}$}}

\confshortname{IDETC/CIE 2015}
\conffullname{the ASME 2015 International Design Engineering Technical Conferences \&\\
              Computers and Information in Engineering Conference}

\confdate{August 2-5}
\confyear{2015}
\confcity{Boston, Massachusetts}
\confcountry{USA}

\papernum{DETC2015-46230}

\title{An Algebraic Method to Check the Singularity-Free Paths for Parallel Robots}

\author{R. Jha, D. Chablat
    \affiliation{
	Institut de Recherche en Communications \\ et 
	Cybern\'etique de Nantes \\
	(UMR CNRS 6597) France\\
    Email addresses: \\ 
		Ranjan.Jha@irccyn.ec-nantes.fr\\
		Damien.Chablat@irccyn.ec-nantes.fr}
}

\author{F. Rouillier     
    \affiliation{
	INRIA Paris-Rocquencourt \\
	Institut de Math\'ematiques de Jussieu \\
	(UMR CNRS 7586),  France \\
	  Email addresses: \\ Fabrice.Rouillier@inria.fr\\}
    }

\author{G. Moroz     
    \affiliation{
	INRIA Nancy-Grand Est, \\
	  France \\
		Email addresses: \\ Guillaume.Moroz@inria.fr}
    }

\begin{document}

\maketitle    

\begin{abstract}
{\it 
Trajectory planning is a critical step while programming the parallel manipulators in a robotic cell. The main problem arises when there exists a singular configuration between the two poses of the end-effectors while discretizing the path with a classical approach. This paper presents an algebraic method to check the feasibility of any given trajectories in the workspace. The solutions of the polynomial equations associated with the trajectories are projected in the joint space using Gr\"{o}bner based elimination methods and the remaining equations are expressed in a parametric form where the articular variables are functions of time $t$ unlike any numerical or discretization method.
These formal computations allow to write the Jacobian of the manipulator as a function of time and to check if its determinant can vanish between two poses. Another benefit of this approach is to use a largest workspace with a more complex shape than a cube, cylinder or sphere. For the Orthoglide, a three degrees of freedom parallel robot, three different trajectories are used to illustrate this method. 
}
\end{abstract}

\section*{INTRODUCTION}
One of the crucial steps in the trajectory planning is to check the singularity-free paths in the workspace for the parallel manipulators. It becomes a necessary protocol to validate the trajectory when the parallel robot is used in practical applications such as precise manufacturing operations. A trajectory verification problem is presented in \cite{merlet:2001} based on some validity criteria like whether the trajectory lies fully inside the workspace of the robot and is singularity-free. Singularity-free path planning for the Gough-Stewart platform with a given starting pose and a given ending pose has been addressed in \cite{das:1998}  using the clustering algorithm is presented in \cite{dash:2003}. An exact method and an approximate method are described in \cite{bhat:1998} to restructure a path close to the singularity locus into a path that avoids it while remaining close to the original path. Due to the geometry of the mechanism, the workspace may not cover fully the space of poses \cite{dash:2003}, hence it is necessary to analyze the workspace of the manipulator.

The workspace of a parallel robot mainly depends on the actuated joint variables, the range of motion of the joints and the mechanical interferences between the bodies of the mechanism. There are different techniques based on geometric tools \cite{Gosselin:1990, Merlet:1992}, discretization \cite{castelli:2008, Ilian:2001, Chablat:2004}, and algebraic methods \cite{zein:2006, ottaviano:2006, Chablat:2014, Chablat:2011} which are used to compute the workspace of a parallel robot. An algebraic method to solve the forward kinematics problem specifically applied to spatial parallel manipulators is described in \cite{luc:2005}. The main advantage of the geometric approach is that, it establishes the nature of the boundary of the workspace \cite{Siciliano:2008}. A procedure to automatically generate the kinematic model of parallel mechanisms which further used for singularity free path planning is reported in \cite{Samir:2008} . 

An algorithm for computing singularity-free paths on closed-chain manipulators is presented in \cite{bohi:2013}, also this method attempts to connect the given two non-singular configurations through a path that maintains a minimum clearance with respect to the singularity locus at all points. The main drawback of any numerical or discretization methods is that there might be a singular configuration between two poses of the end-effectors while discretizing the path. This paper illustrates a technique based on some algebraic methods to check the feasibility of any given trajectories in the workspace : it allows to write the Jacobian of the manipulator as a function of time and to check whether its determinant vanishes between two poses. Also, when the trajectory meets a singularity, its location can also be computed.

The outline of this paper is as follows. We first introduce the modelization and the basic algebraic tools for computing trajectories in the workspace for parallel robots. We then  extended the process to check the singularity-free paths within the workspace and ensures the existence of a singular configurations between the two poses of the end-effector by analysing the Jacobian as a function of time. In later sections we give some detailed examples using three different trajectories for the Orthoglide, a three degrees of freedom parallel robot.

\section*{METHODOLOGY}
\label{secmetho}

To ensure the non existence of a singular configurations between two poses of the end- effector, it is necessary to express the Jacobian of the manipulator as a function of the time. The general procedure to check the feasibility of the trajectories in the workspace as well as a function to compute the projection of the trajectories in the joint space can be decomposed as follows~:
\begin{itemize}
\item 	Defining the constraint equations, articular and pose variables associated with the parallel mechanism;
\item 	Computing the singularities and their projections in the workspace and in the joint space;
\item 	Computing the workspace and joint space boundaries;
\item 	Computing a parametric form of the trajectories in the Cartesian space;
\item 	Computing the projections of the trajectories in the joint space;
\item 	Computing the Jacobian of the manipulator as a function of the time.
\end{itemize}

By computing, we mean, by default, getting a full characterization as solutions of an exact system of algebraic equations. By abuse of language, computing a projection might thus mean computing the algebraic closure of the projection, for example by using classical elimination strategies based on Gr\"{o}bner bases (which thus could introduce a subset of null measure made of spurious points).

Kinematics involves the study of the position, velocity, acceleration, and all higher order derivatives of the position variables (with respect to time or any other parameter/variables). Hence, the study of the kinematics of manipulators refers to all the geometrical and time-based properties of the motion. For a translational manipulator, the set of (algebraic) relations that connects the input values ${\gbf \rho}$ to the output values ${\bf X}$ will be denoted by 
\begin{equation}
    \label{eq1}
         F ({\gbf \rho}, {\bf X}) = 0 
\end{equation}
where ${\gbf \rho}$ is the set of all actuated joint variables and ${\bf X}$ is the set of all pose variables of the end-effector. 

For the study of manipulators, two problems must be considered: the direct kinematic problem (DKP) and the inverse kinematic problem (IKP). Specifically, given a set of joint values, solving the DKP consists in  computing the position and the orientation of the end-effector relatively to the base, or, equivalently to change the representation of the manipulator position from a joint space description into a Cartesian space description. Given the position and orientation of the end-effector of the manipulator,  solving the IKP problem consists in computing all the possible sets of joint angles or parameters that could be used to attain this position and orientation, or, equivalently, mapping locations in three-dimensional Cartesian space to locations in the robot's internal joint space. 

When considering an algebraic modelization, the direct kinematics has basically several solutions, which refers to the several poses of the end-effector for given values of the joint coordinates. It is therefore possible to assemble the manipulator in different ways, and these different configurations are known as the assembly modes of the manipulator \cite{Chablat:1998}. Similarly, the multiple inverse kinematic solutions induce multiple postures for each leg of the manipulator and is termed as the working modes of the manipulator. An analytical relation can be given as ${\bf \rho} = \gamma({\bf X})$ for IKP whereas ${\bf X} = \beta({\bf \rho})$ for DKP. Differentiating Eq.~(\ref{eq1}) with respect to time leads to the velocity model:
 \begin{equation}
    \label{eq2}
         \bf{A} \dot{\bf{X}} + \bf{B} \dot{ \rho} = 0 \quad \quad \quad {\rm where} \quad {\bf A} = \frac{\bf {\partial F}}{\bf {\partial X}} {\rm ,} \quad {\bf B} = \frac{\bf{\partial F}}{{\partial \gbf {\rho}}}
\end{equation}

The matrices \textbf{A} and \textbf{B} are respectively the direct-kinematics and the inverse-kinematics Jacobian matrices of the manipulator. These matrices are used for characterizing different kinds of singularities. The parallel singularities occur whenever ${\rm det}({\bf A})=0$ and the serial singularities occur whenever ${\rm det}({\bf B})=0$. The parallel singular configurations are located inside the workspace. They are particularly undesirable because the manipulator cannot resist to any forces and its control is lost. 

Eliminating $\rho$ in the system $F(\rho,X)=0,~{det}({\bf A})=0$ by means of a Gr\"{o}bner basis computation for a suitable elimination ordering (see \cite{LCO04}) defines (the algebraic closure of) the projection $\xi({\bf X})$ of the parallel singularities in the workspace. In the same way, one can compute (the algebraic closure of) the projection $\varepsilon({\bf X})$ of the parallel singularities in the joint space. Both are then defined as the zero set of some system of algebraic equations and we assume that the considered robots are generic enough so that both are hypersurfaces.

The set of equations associated with the joint limits of the actuator $\chi(\rho)$ can also be projected, with the same elimination method, in the workspace $\mu(\bf X)$ as in Eq.~(\ref{eq:joint_constraints}), where $\rho_{min}$, $\rho_{max}$ are the minimum and maximum values of the articular variables: $\chi(\rho)$ and $\mu(\bf X)$ are the crucial parameters in determining the number of assembly modes and working modes of the parallel manipulator. 

\begin{equation}
\label{eq:joint_constraints}
\chi(\rho) \mapsto \mu(\bf X) \quad \quad  \forall \rho \in (\rho_{min}, \rho_{max}] 
\end{equation}

To use algebraic tools such as Gr\"{o}bner bases, we must represent the trajectories in an algebraic form. A classical approach can be used to transform the trigonometric equations to algebraic ones as in \cite{moroz:2010}. Equation~(\ref{eq:trajectory}) represents the trajectory in parametric form within the workspace as the function of time $t$.

\begin{equation}
    \label{eq:trajectory}
        {\bf X} \mapsto \phi(\bf t)
\end{equation}

$\Psi({\bf X}, \rho, {\bf t})$ in Eq.~(\ref{eq:sys}) is the system of equations which contains kinematic equations $F ({\bf \rho}, {\bf X})$ and the parametric equations of trajectory $\phi(\bf t)$.

\begin{equation}
    \label{eq:sys}
        \Psi({\bf X}, \rho, {\bf t}) = [\phi(\bf t) - {\bf X}, F ({\bf \rho}, {\bf X})]
\end{equation}

This system of equations is projected in the joint space as $\Upsilon({\bf \rho}, {\bf t})$ as shown in Eq.~(\ref{eq:projection}) using Gr\"{o}bner based elimination method.

\begin{equation}
    \label{eq:projection}
        \Psi({\bf X}, \rho, {\bf t}) \mapsto \Upsilon({\bf \rho}, {\bf t})
\end{equation}

The parametric equations of the trajectory as a function of the time in the joint space can then be obtained by solving ${\bf \rho} \gets \Upsilon({\bf \rho}, {\bf t})$.

The workspace analysis allows the characterization of the workspace regions where the number of real solutions for the inverse kinematics is constant. A cylindrical algebraic decomposition (CAD) algorithm is used to compute the workspace of the robot in the projection space ${\bf X}$ with joint constraints $\chi(\rho)$ taken in account \cite{chab:2014, laz:2007, moroz:2008}. 

\noindent The three main steps involved in the analysis are:
\begin{itemize}
\item 	Computation of a subset of the joint space (resp. workspace) where the number of solutions changes: the {\it Discriminant Variety} .
\item 	Description of the complementary of the discriminant variety in connected cells: the Generic {\it Cylindrical Algebraic Decomposition} (CAD).
\item 	Connection of the cells belonging to the same connected component in the counterpart of the discriminant variety: {\it interval comparisons.}
\end{itemize}

The joint space analysis predicts the feasible and non-feasible combinations of the prismatic joint variables which are essential for the parallel robot control. The joint space analysis allows the characterization of the regions where the number of real solutions for the direct kinematic problem is constant. The joint space analysis is done using CAD, which gives the number of cells corresponding to the number of solutions in the joint space. $\chi(\rho)$ parameter significantly changes the number of assembly modes and working modes of the manipulator.

Substituting $\phi(\bf t)$ in $\xi(\bf X)$ (in the polynomial defining $\xi(\bf X)$ ) we then get $\xi(\phi(\bf t))$, which defines the vanishing values of the Jacobian as a function of the time.  This equation in $cos(t),sin(t)$ can be turned into a zero-dimensional bivariate system and its solutions can thus be computed exactly, either by means of a rational parametrization or by means of isolating intervals with rational bounds that can be refined to any arbitrary precision \footnote{for example the functions Groebner[RationalUnivariateRepresentation] and RootFinding[Isolate] in {\em Maple} software}.

Whatever the chosen (exact) representation for a solution, it can easily be checked if it vanishes between the two poses of the end-effector. The solutions of  $\xi(\bf \phi(\bf t))$contain the singular points on the trajectory $\phi(\bf t)$ and eventually few spurious points due to the projection of ${\rm det}({\bf A})$ in Cartesian space. Spurious singular points can then be differentiated from real singular points by substituting $\phi(\bf t)$ and the solutions of $\Upsilon({\bf \rho}, {\bf t})$ in ${\rm det}({\bf A})$.

\section*{METHOD VALIDATION}
We propose to validate our approach (and related tools) by checking the feasibility of three different trajectories for the Othoglide. The first two trajectories are heart shaped planar parametric curves and the third trajectory is a parametric curve in three dimensions. 

These trajectories are fully inside the workspace and are selected such that their projections in the joint space is defined by parametric equations containing trigonometric functions. In addition, we have shown that the singular and singular-free trajectories are well discriminated, as one of the trajectory cuts the singularity surface.
\subsection*{Manipulator Architecture and Kinematics}
\begin{figure}
    \begin{center}
    \begin{tabular}{@{}c@{}c@{}}
       \begin{minipage}[t]{46 mm}
				\psfrag{p1}{$P1$}
				\psfrag{p2}{$P2$}
				\psfrag{a1}{$A1$}
				\psfrag{a2}{$A2$}
				\psfrag{a3}{$A3$}
				\psfrag{b1}{$B1$}
				\psfrag{b2}{$B2$}
				\psfrag{b3}{$B3$}
				\includegraphics[width=43 mm]{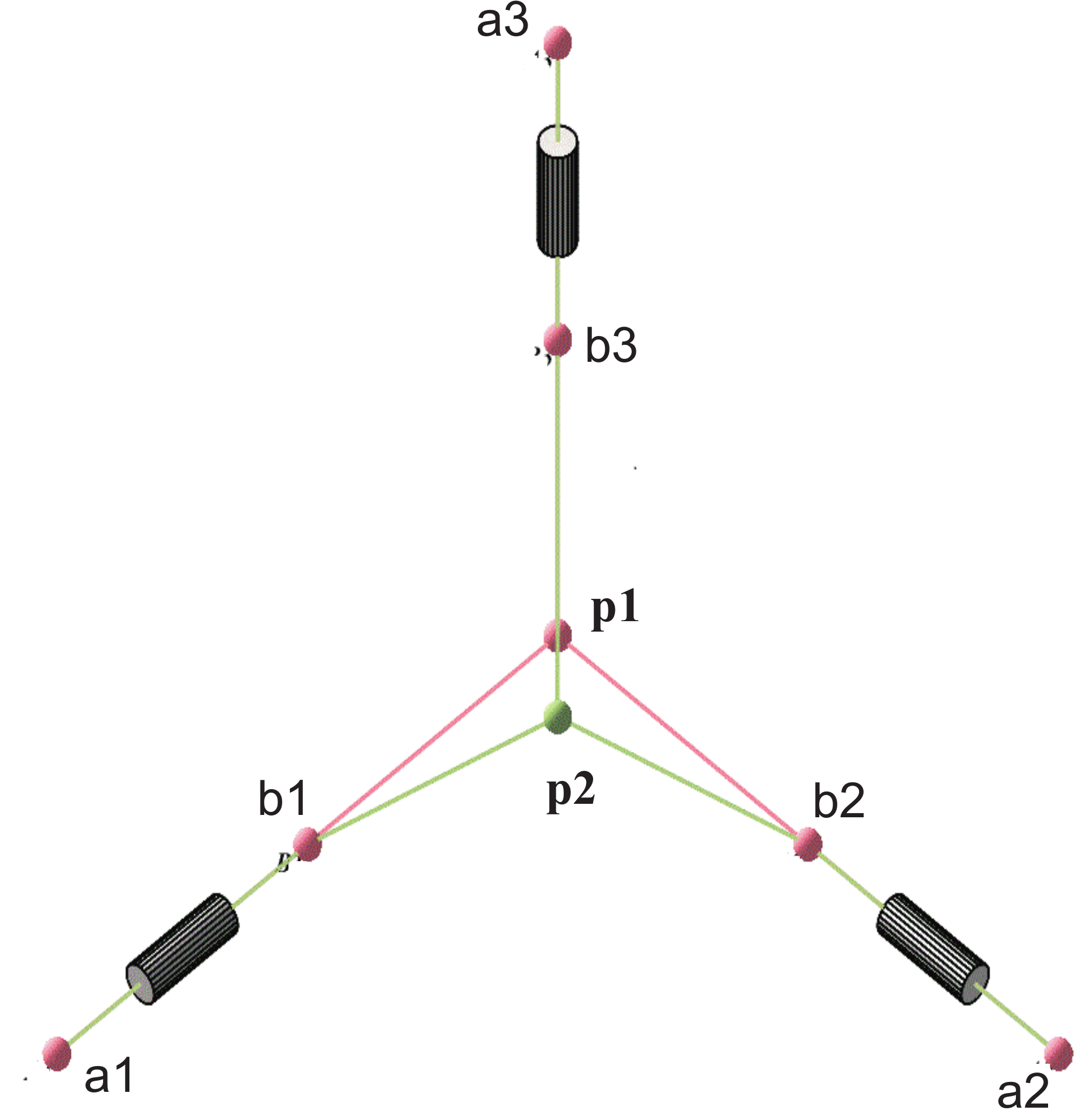}
       \end{minipage} &
       \begin{minipage}[t]{46 mm}
				\psfrag{z}{$z$}
				\psfrag{x}{$x$}
				\psfrag{y}{$y$}
				\psfrag{ik}{$1$ IKP}
				\psfrag{dk}{$8$ IKP}
				\includegraphics[width=43 mm]{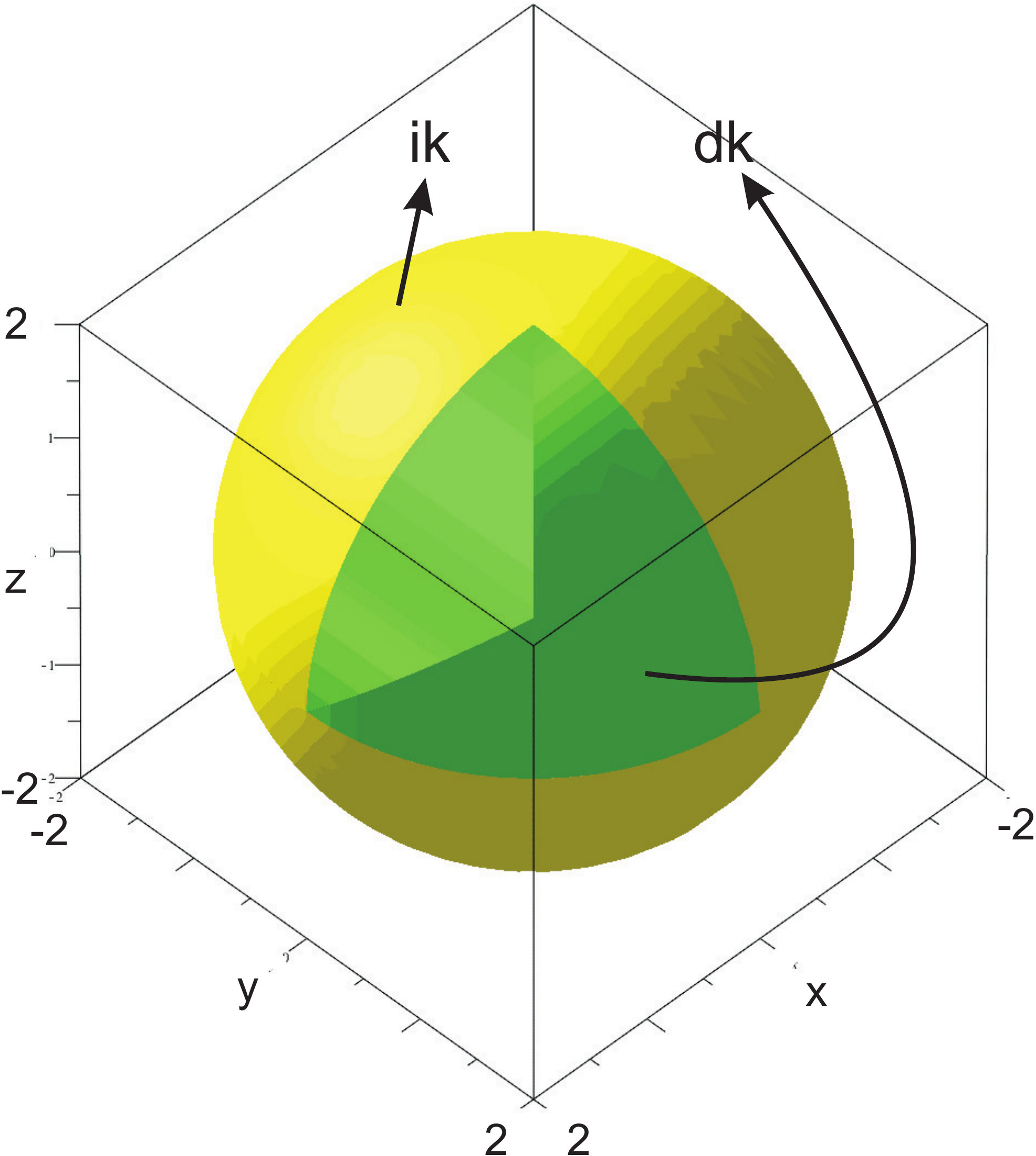}
       \end{minipage} \\
			{\small (a)} & {\small (b)} \\
       \begin{minipage}[t]{46 mm}
				\psfrag{z}{$z$}
				\psfrag{x}{$x$}
				\psfrag{y}{$y$}
				\psfrag{si}{$\xi(\bf X)$}
				\includegraphics[width=43 mm]{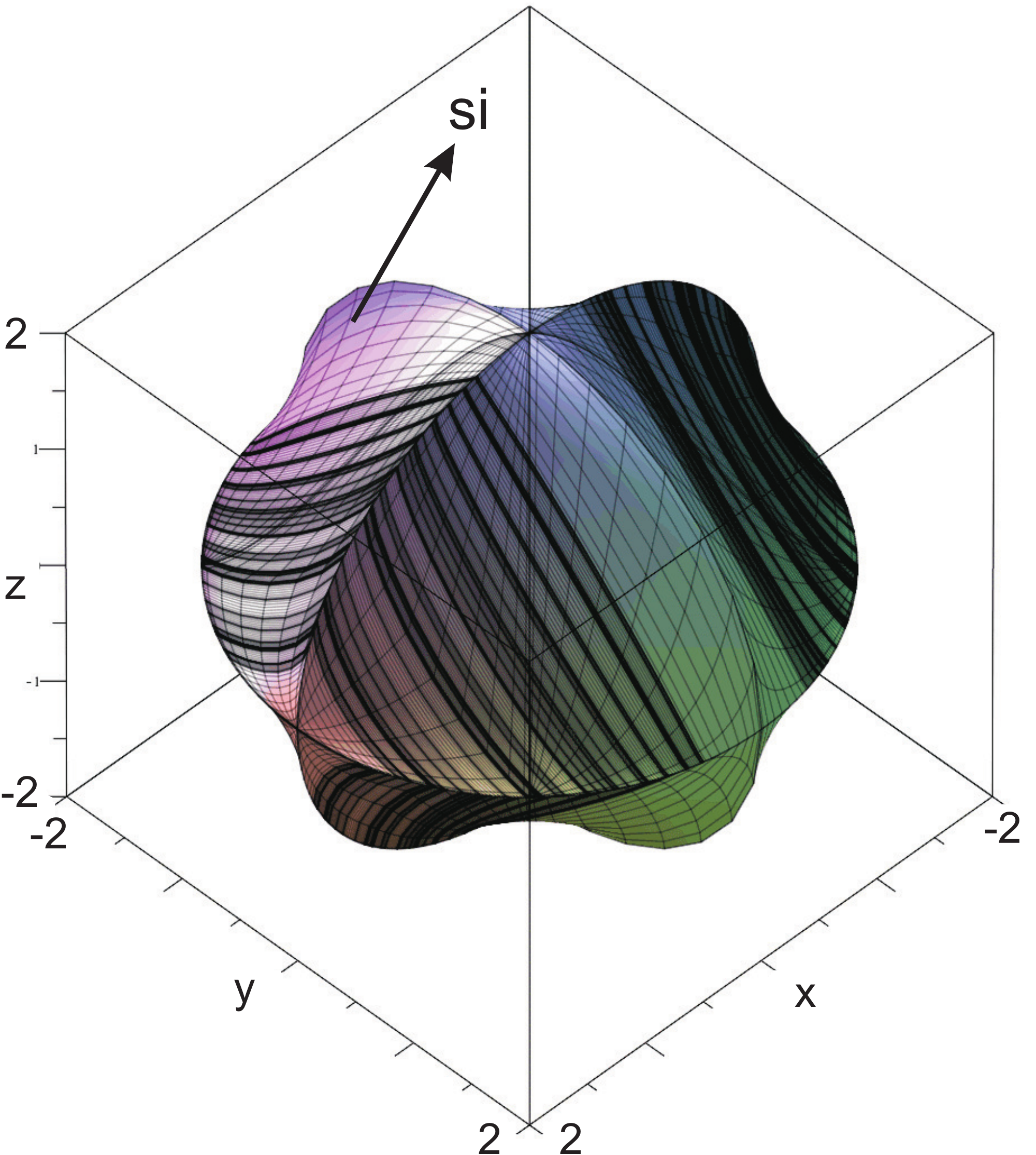}
       \end{minipage} &
       \begin{minipage}[t]{46 mm}
				\psfrag{r1}{$\rho_1$}
				\psfrag{r2}{$\rho_2$}
				\psfrag{r3}{$\rho_3$}
				\psfrag{js}{$2$ DKP}
				\includegraphics[width=43 mm]{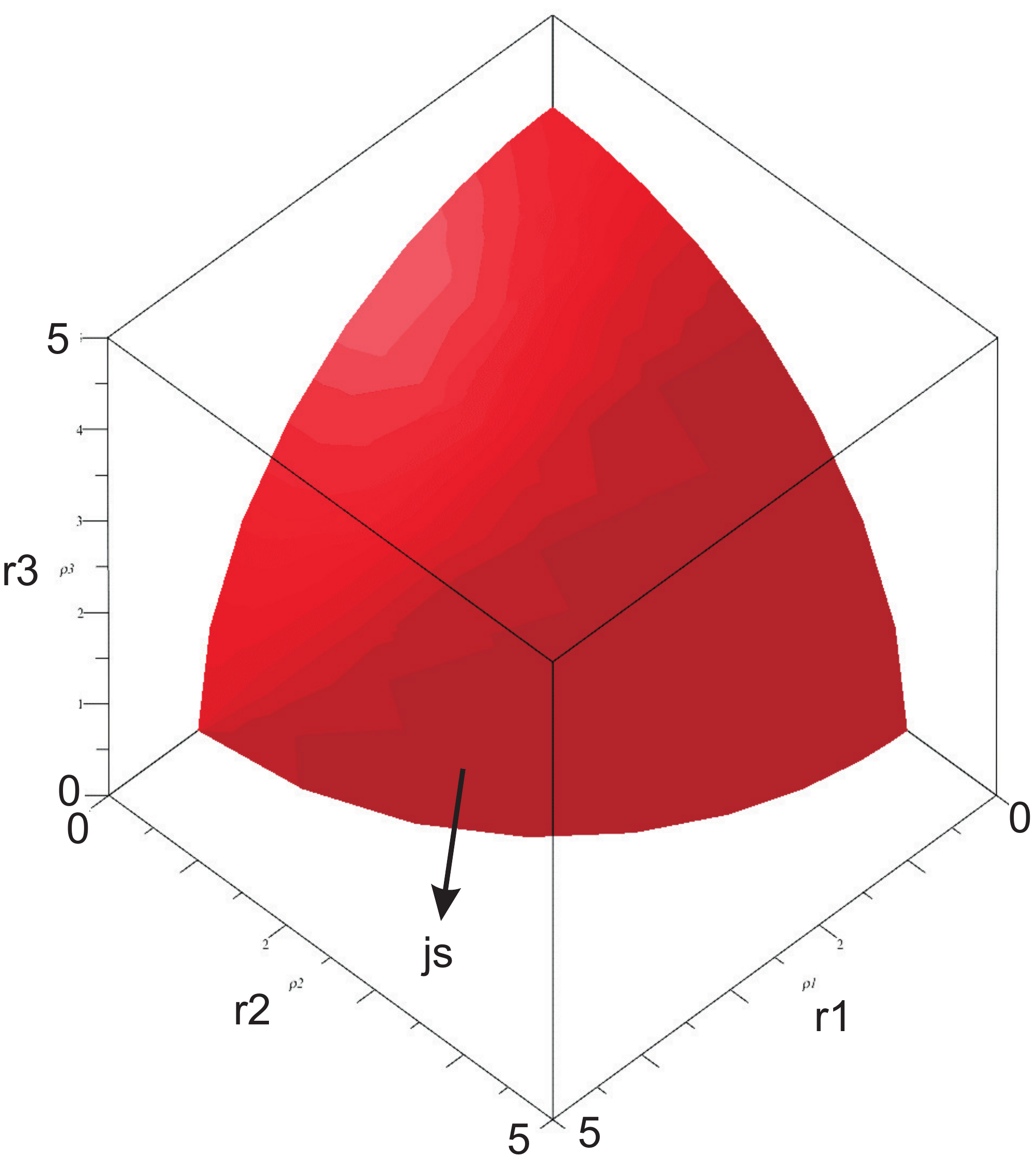}
       \end{minipage} \\
			{\small (c)} & {\small (d)} 
    \end{tabular}
    \end{center}
    \caption{Architecture of an Orthoglide including two assembly modes (a) Workspace plot with $\chi(\rho)$ joint constraints (b) and Parallel singularity ${\rm det}({\bf A})$ projected in the workspace as $\xi(\bf X)$ (c) and joint space plot with $\chi(\rho)$ joint limits (d)}
    \protect\label{figure:architecture}
\end{figure}

The manipulator under the study is an {\bf Orthoglide} parallel robot with three degrees of freedom. The mechanism is driven by three actuated orthogonal prismatic joints $\rho_i$, made of three parallelogram connected by revolute joints to the tool center point on one side and to the corresponding prismatic joint at another side. The assembly modes of these robots depends on the solutions of the DKP as shown in Fig.~\ref{figure:architecture}(a). The point ${\bf P}_i$ represents the pose of corresponding robot. However more than one value of $i$ for the point ${\bf P}_i$ denote multiple solutions for the DKP. ${{\bf A}_i}{\bf B}_i$ is equal to $\rho_i$, where ${\bf \rho}_i$ represents the prismatic joint variables whereas ${\bf P}$ represents the position vector of the tool center point.

The constraint equations for the Orthoglide with the actuated variables $\rho = [\rho_1, \rho_2, \rho_3]$  and the pose variables ${\bf X} = [x, y, z]$ are:

\begin{eqnarray}\label{eq:Orthoglide}
F ({\bf \rho}, {\bf X}): (x-\rho_1)^2 + y^2 + z^2  = l^2 \nonumber \\ 
 x^2 + (y-\rho_2)^2 + z^2  = l^2 \nonumber \\
 x^2 + y^2 + (z-\rho_3)^2  = l^2 
\end{eqnarray}

\noindent where $l$ is the leg length of the manipulator and is equal to two for all the computations. The set of equations associated with the joint limits of the actuator $\chi(\rho)$, projected in the workspace $\mu(\bf X)$, are given in Eq. (\ref{eq:cnstrnt}). $\chi(\rho)$ plays an important role in determining the shape of the workspace and singularity surfaces. It also affects the number of solutions for the IKP {\it i.e.} the working modes associated with the manipulator. Figure~\ref{figure:architecture}(b) represents the workspace of the Orthoglide. A CAD algorithm is used to compute the workspace of the robot in the projection space $(x, y, z)$, taking into account the joint constraints $\chi(\rho)$ . Without considering the joint limits, the Orthoglide admits two assembly modes and eight working modes \cite{Pashkevich:2006}.

\begin{eqnarray}
\label{eq:cnstrnt}
\chi(\rho) = [\rho_1,\rho_2,\rho_3,-4+\rho_1,-4+\rho_2,-4+\rho_3] \nonumber \\
\chi(\rho) \mapsto \mu(\bf X)  \nonumber \\ 
{\rho_i} \mapsto { x^2} + { y^2} + { z^2} - 4 \quad {i=1,2,3}\nonumber \\
 -4 + \rho_i \mapsto { x^2} + { y^2} + { z ^2} - 8{ x} + 12 \nonumber \\
 -4 + \rho_i \mapsto { x^2} + { y^2} + { z ^2} - 8{ y} + 12 \nonumber \\
 -4 + \rho_i \mapsto { x^2} + { y^2} + { z ^2} - 8{ z} + 12
\end{eqnarray}

\begin{figure}
    \begin{center}
    \begin{tabular}{@{}c@{}c@{}}
       \begin{minipage}[t]{48 mm}
				\psfrag{z}{$z$}
				\psfrag{x}{$x$}
				\psfrag{y}{$y$}
				\psfrag{si}{$\xi(\bf X)+\mu(\bf X)$}
				\includegraphics[width=45mm]{Figures/Orthoglide_ps-eps-converted-to.pdf} 
       \end{minipage} &
       \begin{minipage}[t]{48 mm}
				\psfrag{z}{$z$}
				\psfrag{x}{$x$}
				\psfrag{y}{$y$}
				\psfrag{sim}{$\xi(\bf X)$}
				\includegraphics[width=45mm]{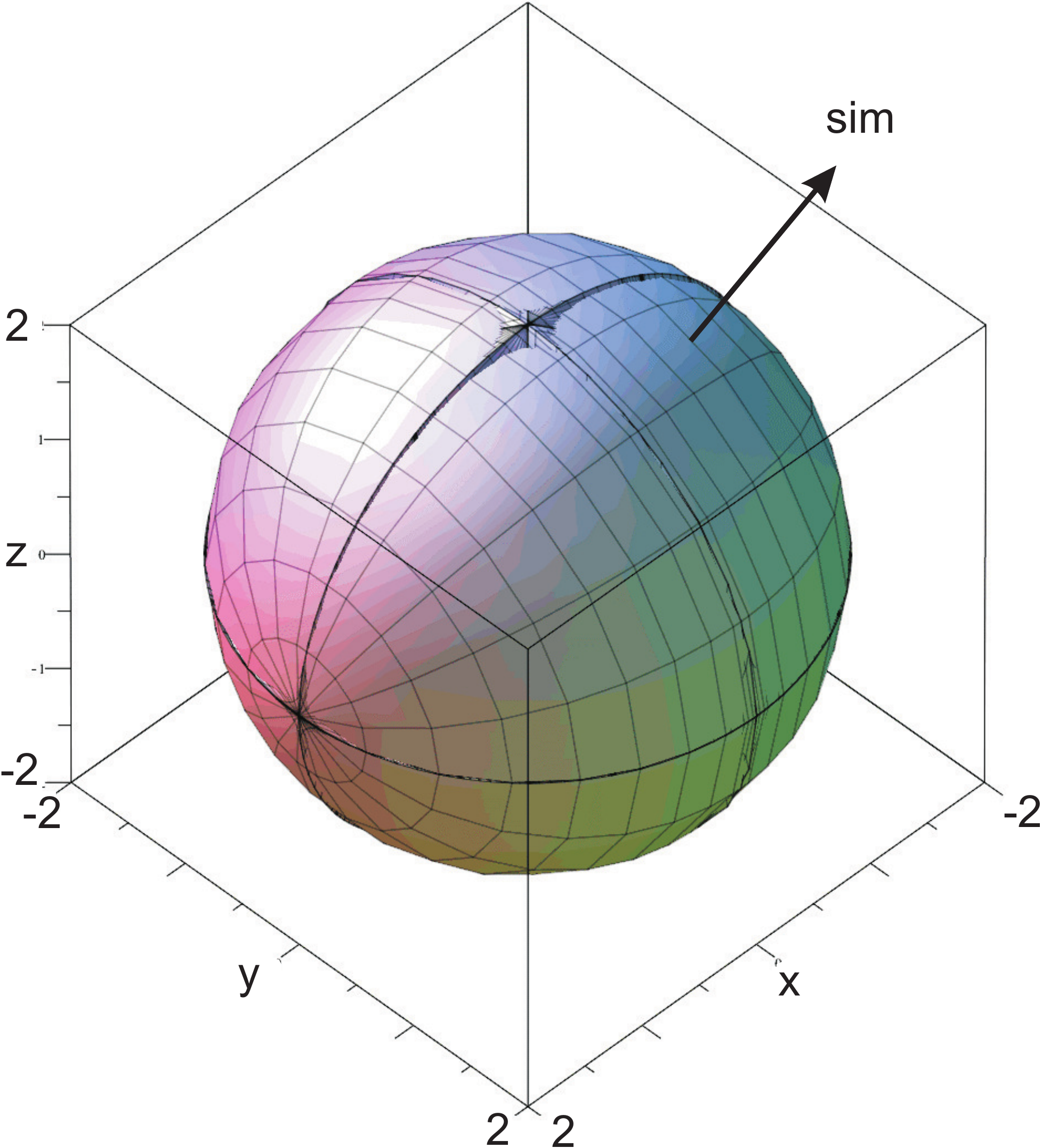} 
       \end{minipage} \\
				{\small (a)} & {\small (b)}
    \end{tabular}
    \caption{A comparison between the parallel singularity surface for an Orthoglide computed with the joint limits $\mu(\bf X)$ (a) without $\mu(\bf X)$ (b) }
    \protect\label{figure:sing_cnstr}
    \end{center}
\end{figure}

In the Figure ~\ref{figure:sing_cnstr}(b), the yellow region (resp. green region), corresponds to the region where the inverse kinematic model has two real solutions (resp. height).

The joint space analysis is done using CAD to find the regions where the direct kinematics problem admits a fixed number of real solutions. For example, in Fig.~\ref{figure:architecture}(d), the red cell corresponds to to the region where the DKP has two solutions. 

There is a difference in the shape of the singularity surface, depending if we consider the joint constraints $\mu(\bf X)$ or not (Fig.~\ref{figure:sing_cnstr}).

\begin{eqnarray}
\label{eq:singlr}
{{\rm det}({\bf A})} = -8\rho_1\rho_2\rho_3 + 8\rho_1\rho_2{z} + 8\rho_1\rho_3{ y} + 8\rho_2\rho_3{ x} \nonumber \\ 
{\rm det}({\bf A}) \mapsto \varepsilon(\bf \rho)  \quad \quad {\rm det}({\bf A}) \mapsto \xi(\bf X) \nonumber \\ 
{\varepsilon(\bf \rho)} = \rho_1^4\rho_2^2+\rho_1^4\rho_3^2+\rho_1^2\rho_2^4+2\rho_1^2\rho_2^2\rho_3^2+\rho_1^2\rho_3^4+ \nonumber \\\rho_2^4\rho_3^2+\rho_2^2\rho_3^4-16\rho_1^2\rho_2^2-16\rho_1^2\rho_3^2-16\rho_2^2\rho_3^2 
\end{eqnarray}

In Eq. (\ref{eq:singlr}), ${\rm det}({\bf A})$ is the parallel singularity of the Orthoglide and $\xi(\bf X)$ is the projection of ${\rm det}({\bf A})$ in workspace. The mathematical expression for $\xi(\bf X)$ is not displayed in Eq. (\ref{eq:singlr}) due to the lack of space.  Fig.~\ref{figure:architecture}(c) shows the projection $\xi(\bf X)$ which is plotted with $\mu(\bf X)$ as one of the input parameter. The degree of this characteristic surface is $18$ and it represents the singularities associated with the eight working modes.

\subsection*{Trajectory definition in the workspace}

\begin{figure}
    \begin{center}
    \begin{tabular}{@{}c@{}c@{}}
       \begin{minipage}[t]{49 mm}
				\psfrag{z}{$z$}
				\psfrag{x}{$x$}
				\psfrag{y}{$y$}
				\psfrag{trajectory}{Trajectory 1}
				\psfrag{trajecto}{Trajectory 2}
				\psfrag{wrk}{$Workspace$}
				\includegraphics[width=45mm]{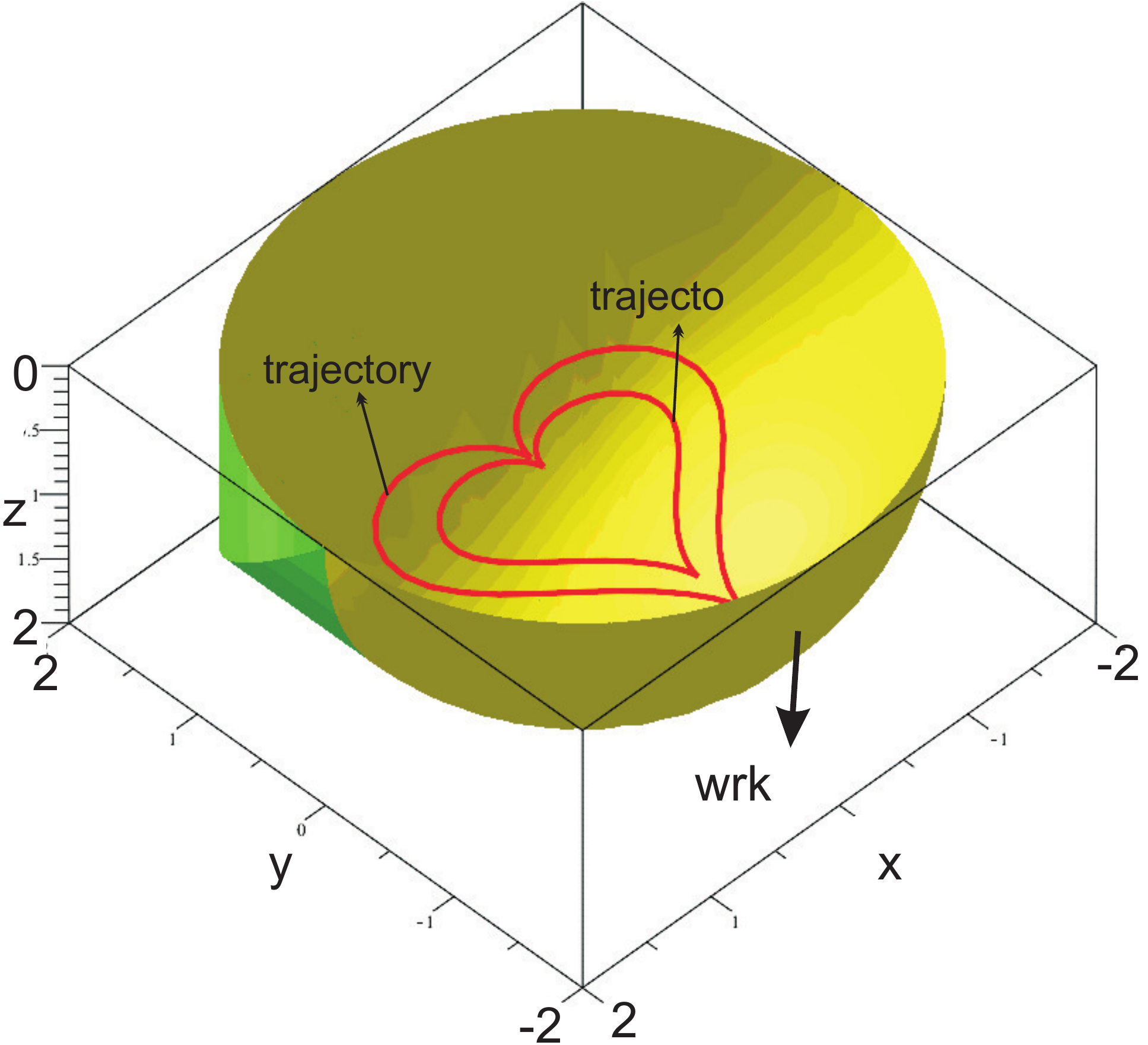} 
       \end{minipage} &
       \begin{minipage}[t]{49 mm}
				\psfrag{z}{$z$}
				\psfrag{x}{$x$}
				\psfrag{y}{$y$}
				\psfrag{traj}{Trajectory 3}
				\includegraphics[width=45mm]{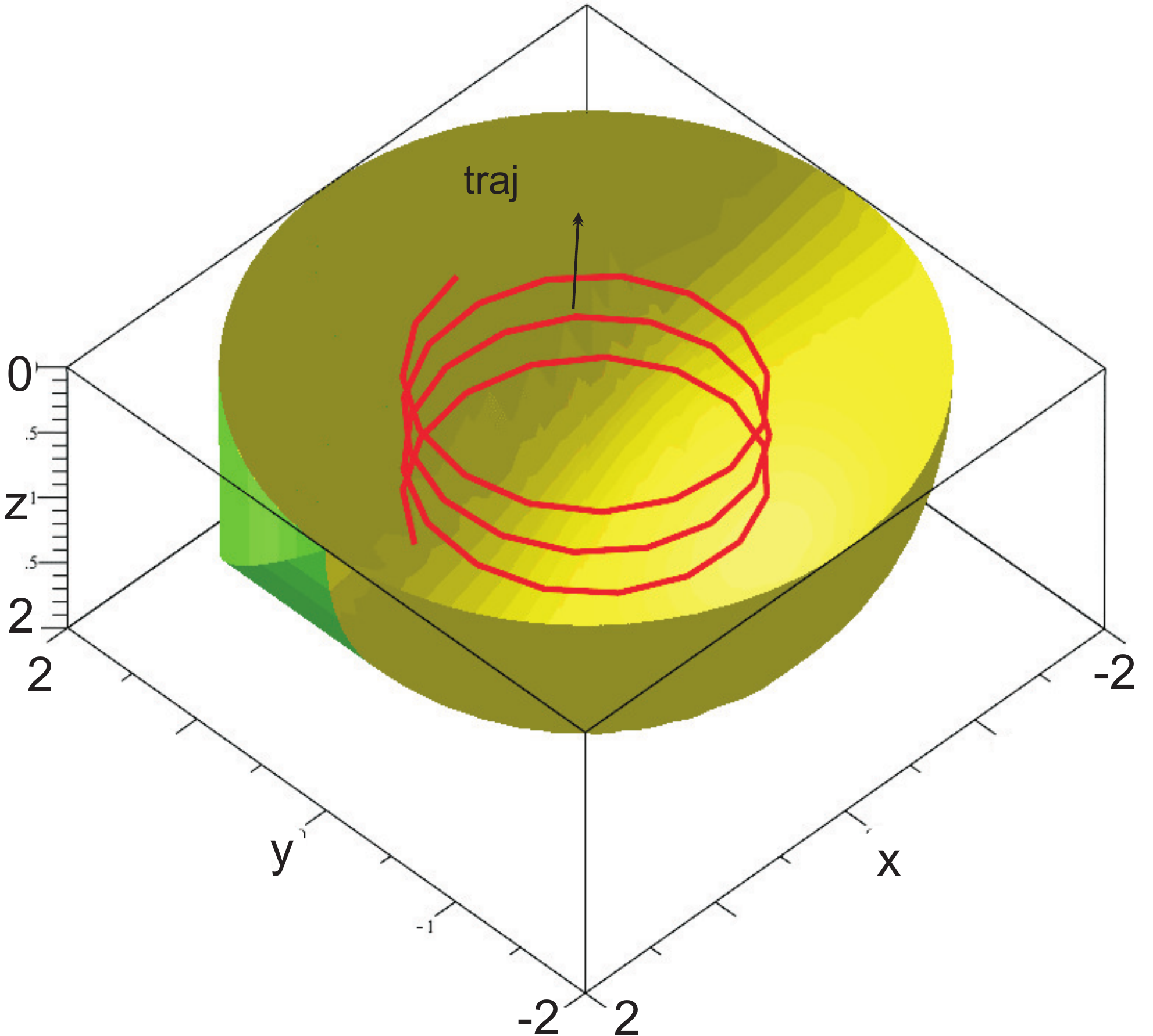} 
       \end{minipage} \\
				{\small (a)} & {\small (b)}
    \end{tabular}
    \caption{Position of Trajectories 1\&2 (a) Trajectory 3 (b) in the workspace of an Orthoglide}
    \protect\label{figure:Trajectory}
    \end{center}
\end{figure}

Trajectories 1 \& 2 are heart shaped parametric curves which ( Fig.~\ref{figure:Trajectory}(a)). Eq. (\ref{eq:Trajectory1}) and Eq. (\ref{eq:Trajectory2}) are the mathematical definitions of  Trajectory 1 and Trajectory 2, respectively. As these equations are trigonometric equations, it is necessary to represent them in an algebraic form. $\phi_1(t)$ and $\phi_2(t)$ are the parametric equations for Trajectory 1 and Trajectory 2, respectively, which are defined for $t \in [-\pi, \pi]$. From now, $\sin$ and $\cos$ are replaced by ${\bf s}$ and ${\bf c}$ respectively to reduce the size of the expressions.

\begin{eqnarray}\label{eq:Trajectory1}
 \phi_1(t): \quad \quad \quad \quad x = \frac{8}{7}{\bf s}^3(t) \nonumber \\ 
 y = \frac{13}{14}{\bf c}(t)-\frac{5}{14}{\bf c}(2t)-\frac{1}{10}{\bf c}(3t)-\frac{1}{14}{\bf c}(4t) \nonumber \\
 z = 1 
\end{eqnarray}

\begin{eqnarray}\label{eq:Trajectory2}
 \phi_2(t): \quad \quad \quad \quad x =  \frac{4}{5}{\bf s}^3(t) \nonumber \\ 
 y = \frac{13}{20}{\bf c}(t)-\frac{1}{4}{\bf c}(2t)-\frac{1}{20}{\bf c}(3t)-\frac{1}{20}{\bf c}(4t) \nonumber \\
 z = 1 
\end{eqnarray}

Trajectory 3 is a parametric curve in three dimensions. Fig.~\ref{figure:Trajectory}(b) represents the trajectory in the workspace of the manipulator. The parametric equations of the trajectory is defined by Eq. (\ref{eq:Trajectory1}). $\phi_3(t)$  is the parametric equation for Trajectory 3  and is defined for $t \in [0,\ 20]$. 

\begin{eqnarray}\label{eq:Trajectory3}
 \phi_3(t): \quad x = {\bf s}(t) \quad y = {\bf c}(t) \quad z = \frac{1}{20}t 
\end{eqnarray}

In order to turn the system to an algebraic one, we add the following equations
\begin{eqnarray}\label{eq:assum}
 \sin(t)= {\sin\_t} \quad \cos(t)= {\cos\_t}  \nonumber \\ 
 \sin\_t^2 + \cos\_t^2 = 1
\end{eqnarray}
\noindent and we remark that this change of variables does not introduce any spurious solutions since it is bjective ($t\in[-\pi,\pi]$).

\subsection*{Projection in the joint space}
\begin{figure*}
    \begin{center}
        \psfrag{z}{$z$}
				\psfrag{x}{$x$}
				\psfrag{y}{$y$}
				\psfrag{r1}{$\rho_1$}
				\psfrag{r2}{$\rho_2$}
				\psfrag{r3}{$\rho_3$}
				\psfrag{tr}{Trajectory 1}
				\psfrag{tra}{Trajectory 2}
				\psfrag{work}{WORKSPACE}
				\psfrag{je}{JOINTSPACE}
				\psfrag{Projection}{Projection}
				\psfrag{corr}{Corresponds to 8 IKP}
				\includegraphics[width=140mm]{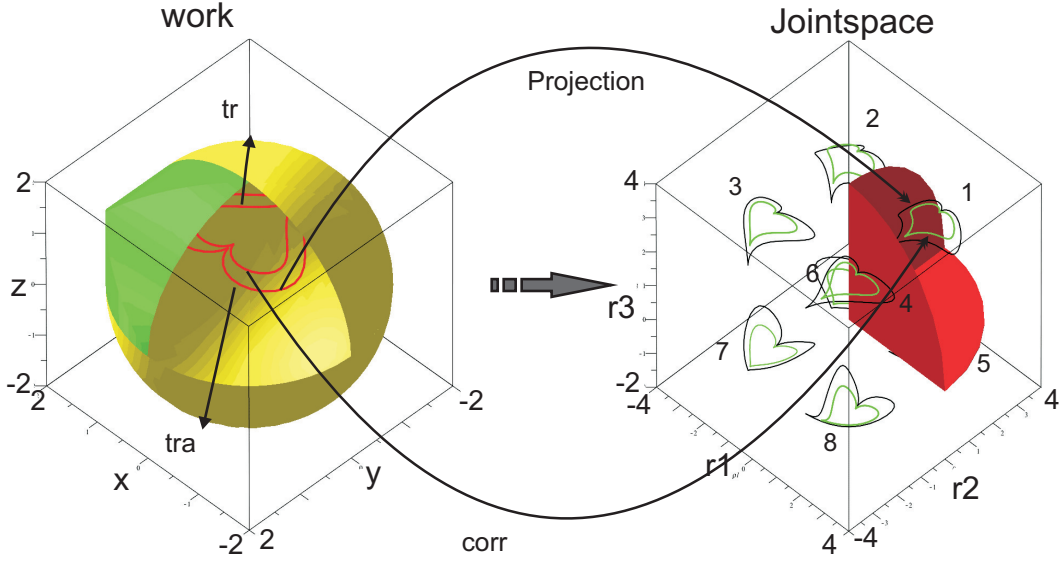} 
       \caption{A pictorial representation of the mapping of trajectories from workspace to joint space. Eight different pairs of trajectories, $\Upsilon_1({\bf \rho}, {\bf t})$ \& $\Upsilon_2({\bf \rho}, {\bf t})$ in joint space are the image of corresponding $\phi_1(\bf t)$ \& $\phi_2(\bf t)$. These eight different trajectories are associated with the eight working modes of the Orthoglide. Only one trajectory lies inside the joint space boundary due to to the joint constraints.}
    \protect\label{figure:projection_traj12}
    \end{center}
\end{figure*}

$\phi_1(t)$, $\phi_2(t)$ and $\phi_3(t)$ are the trajectories which are defined in the workspace. To project these trajectories in the joint space, it is necessary to formulate a system of equations corresponding to each trajectory which also consists the kinematic equations of the manipulator. ${\Psi_1}$, ${\Psi_2}$ and ${\Psi_3}$ are the corresponding systems of equations for the  Trajectories 1, 2 and 3, respectively (see Eq. (\ref{eq:sysf})).

\begin{eqnarray}
    \label{eq:sysf}
        {\Psi_1}({\bf X}, \rho, {\bf t}) = [\phi_1(\bf t) - {\bf X}, F ({\bf \rho}, {\bf X})] \nonumber \\ 
				{\Psi_2}({\bf X}, \rho, {\bf t}) = [\phi_2(\bf t) - {\bf X}, F ({\bf \rho}, {\bf X})]\nonumber \\ 
				{\Psi_3}({\bf X}, \rho, {\bf t}) = [\phi_3(\bf t) - {\bf X}, F ({\bf \rho}, {\bf X})]
\end{eqnarray}

\begin{figure}
    \begin{center}
    \begin{tabular}{@{}c@{}c@{}}
       \begin{minipage}[t]{73 mm}
				\psfrag{r1}{$\rho_1$}
				\psfrag{r2}{$\rho_2$}
				\psfrag{r3}{$\rho_3$}
				\psfrag{jp}{$joint space$}
				\psfrag{bp}{$Boundary$}
				\psfrag{fea}{$Feasible \quad Trajectory$}
				\psfrag{Trajectory 3}{$Trajectory 3$}
				\includegraphics[width=68mm]{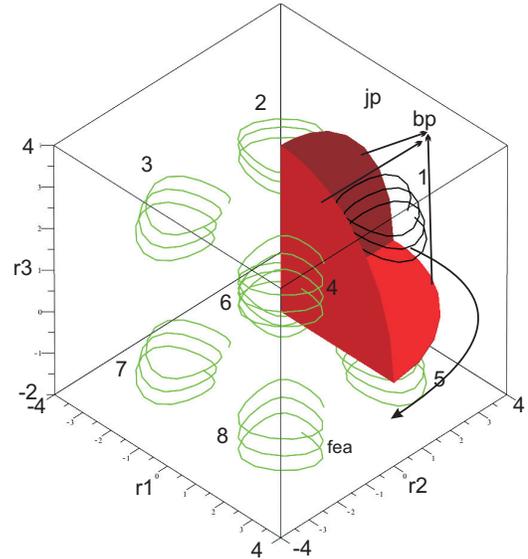} 
       \end{minipage} 
			\end{tabular}
    \caption{Mapping of the Trajectory 3 from workspace to joint space. Eight different possible solutions of $\Upsilon_3({\bf \rho}, {\bf t})$  are marked in joint space which are the image of $\phi_3(\bf t)$. There is only one feasible trajectory (marked as $1$) lies inside the joint space boundary.}
    \protect\label{figure:projection_traj3}
    \end{center}
\end{figure}

Each system of is projected in the joint space as $\Upsilon_1({\bf \rho}, {\bf t})$, $\Upsilon_2({\bf \rho}, {\bf t})$ and $\Upsilon_3({\bf \rho}, {\bf t})$ (see  Eq.~(\ref{eq:sng1})). By solving $\Upsilon_1({\bf \rho})$, $\Upsilon_2({\bf \rho})$ and $\Upsilon_3({\bf \rho})$ for $\rho$, we get the corresponding parametric equations for the trajectories in the joint space, as shown in Fig.~\ref{figure:projection_traj12} for the Trajectories $1$\&$2$ and in Fig.~\ref{figure:projection_traj3} for the Trajectory 3. Eqs.~(\ref{eq:Projection3})~-~(\ref{eq:Projection2}) are used to plot all possible images of the trajectories in the joint space. All the computations are done without considering the joint limits. As the Orthoglide has eight working mode, there exists eight possible trajectories in  the joint space ( Fig.~\ref{figure:projection_traj12} and Fig.~\ref{figure:projection_traj3} for Trajectories 1\&2 and Trajectory 3, respectively).

\begin{eqnarray}
\label{eq:sng1}
  \Psi_1({\bf X}, \rho, {\bf t}) \mapsto \Upsilon_1({\bf \rho}, {\bf t}) \quad \quad \forall t \in [-\pi, \pi]\nonumber \\ 
  \Psi_2({\bf X}, \rho, {\bf t}) \mapsto \Upsilon_2({\bf \rho}, {\bf t}) \quad \quad \forall t \in [-\pi, \pi]\nonumber \\
  \Psi_3({\bf X}, \rho, {\bf t}) \mapsto \Upsilon_3({\bf \rho}, {\bf t}) \quad \quad \forall t \in [0,~\ 20]
  \end{eqnarray}

Solving $\Upsilon_1({\bf \rho}, {\bf t})$ and $\Upsilon_2({\bf \rho}, {\bf t})$ for $\rho$ gives eight possible solutions ( Eq.~(\ref{eq:Projection1}) for Trajectory 1 and  Eq.~(\ref{eq:Projection2}) for Trajectory 2), which inferred eight different possible images of Trajectory 1\&2 in joint space. But from Fig.~\ref{figure:projection_traj12} one can see that only one trajectory lies inside the joint space boundary for the both trajectories in workspace. 

\begin{eqnarray}
\label{eq:Projection3}
  \rho_1 = {\bf s}(t) \pm \frac{1}{20}\sqrt{400{\bf s}^2(t)-t^2+1200} \nonumber \\ 
 \rho_2 = {\bf c}(t) \pm \frac{1}{20}\sqrt{400{\bf c}^2(t)-t^2+1200} \nonumber \\
 \rho_3 = \frac{1}{20}t \pm \sqrt{3}
  \end{eqnarray}
	
	Figure~\ref{figure:projection_traj3} shows all the possible images of Trajectory 3 in the joint space. Solving $\Upsilon_3({\bf \rho}, {\bf t})$ for $\rho$ gives eight possible solutions which (Eq.~(\ref{eq:Projection3})). It can be seen in the Fig.~\ref{figure:projection_traj3} that only one trajectory (which is marked as 1), lies inside the joint space boundary.

\begin{figure*}
\begin{align}\label{eq:Projection1}
\rho_1 &= \frac{8}{7}{\bf s}^3(t) \pm \frac{1}{35}\sqrt{1996{\bf s}^6(t)-400{\bf s}^8(t)+560{\bf s}^6(t) {\bf c}(t)-100{\bf s}^4(t){\bf c}(t)-2009{\bf s}^4(t)+2190{\bf c}^3(t)-1379{\bf c}^2(t)-1320{\bf c}(t)+3988} \nonumber \\ 
 \rho_2 &= -\frac{4}{7}{\bf c}^4(t)-\frac{2}{5}{\bf c}^3(t)-\frac{1}{7}{\bf c}^2(t)+\frac{43}{35}{\bf c}(t)+\frac{2}{7} \pm \frac{1}{7}\sqrt{64{\bf c}^6(t)-192{\bf c}^4(t)+192{\bf c}^2(t)+83} \nonumber \\
 \rho_3 &= 1 \pm \frac{1}{35}\sqrt{3200-400{\bf c}^8(t)-560{\bf c}^7(t)+1204{\bf c}^6(t)+1580{\bf c}^5(t)-3221{\bf c}^4(t)+710{\bf c}^3(t)+3051{\bf c}^2(t)-860{\bf c}(t)}
\end{align}
\end{figure*}

\begin{figure*}
\begin{align}\label{eq:Projection2}
\rho_1 &= \frac{4}{5}{\bf s}^3(t) \pm \frac{1}{10}\sqrt{76{\bf s}^6(t)-16{\bf s}^8(t)+16{\bf s}^6(t){\bf c}(t)+12{\bf s}^4(t){\bf c}(t)-85{\bf s}^4(t)+96{\bf c}^3(t)-66{\bf c}^2(t)-60{\bf c}(t)+321} \nonumber \\ 
 \rho_2 &= -\frac{2}{5}{\bf c}^4(t)-\frac{1}{5}{\bf c}^3(t)-\frac{1}{10}{\bf c}^2(t)+\frac{4}{5}{\bf c}(t)+\frac{1}{5} \pm \frac{1}{5}\sqrt{16{\bf c}^6(t)-48{\bf c}^4(t)+48{\bf c}^2(t)+59} \nonumber \\
 \rho_3 &= 1 \pm \frac{1}{10}\sqrt{332-16{\bf c}^8(t)-16{\bf c}^7(t)^7+52{\bf c}^6(t)+60{\bf c}^5(t)-145{\bf c}^4(t)+24{\bf c}^3(t)+132{\bf c}^2(t)-32{\bf c}(t)}
\end{align}
\end{figure*}

\subsection*{Singularity analysis}
To check if there exists any singular configuration between the two poses of the  end-effector it is necessary to express the Jacobian of the manipulator as a function of the time or of some independent variable. The proposed algebraic method enables us to write the Jacobian of the manipulator as a function of the time and to check if its determinant vanishes between two poses. 

\begin{figure}
    \begin{center}
    \begin{tabular}{@{}c@{}c@{}}
       \begin{minipage}[t]{72 mm}
				\psfrag{s1}{$s_1$}
				\psfrag{s2}{$s_2$}
				\psfrag{s3}{$s_3$}
				\psfrag{s4}{$s_4$}
				\psfrag{t}{$t$}
				\psfrag{det(j)}{$\mu_1(\bf t)$}
				\psfrag{p}{$-\pi$}
				\psfrag{-p}{$\pi$}
				\psfrag{spurs}{Spurious}
				\psfrag{singu}{Singular Points}
				\includegraphics[width=68mm]{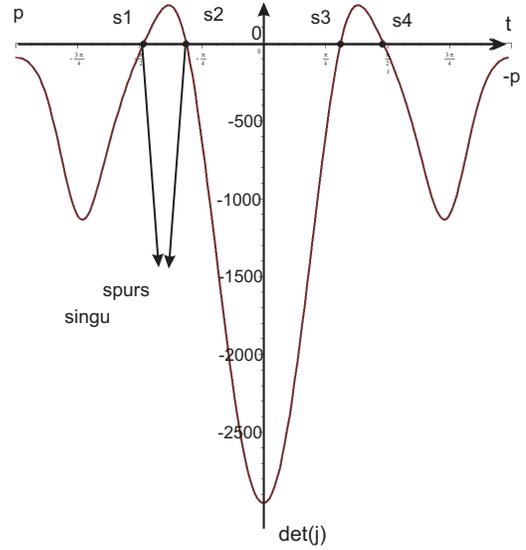} 
       \end{minipage} 
			\end{tabular}
    \end{center}
		\caption{Variation of $\mu_1(\bf t)$ along Trajectory 1. $s_1$, $s_2$, $s_3$ and $s_4$ represents the four solutions of $\mu_1(\bf t) = 0 \quad \forall t \in [-\pi, \pi]$.}
		\protect\label{figure:sing11}
\end{figure}

\begin{figure}
    \begin{center}
    \begin{tabular}{@{}c@{}c@{}}
       \begin{minipage}[t]{70 mm}
				\psfrag{t}{$t$}
				\psfrag{det(j)}{$\mu_2(\bf t)$}
				\psfrag{p}{$-\pi$}
				\psfrag{-p}{$\pi$}
				\includegraphics[width=67mm]{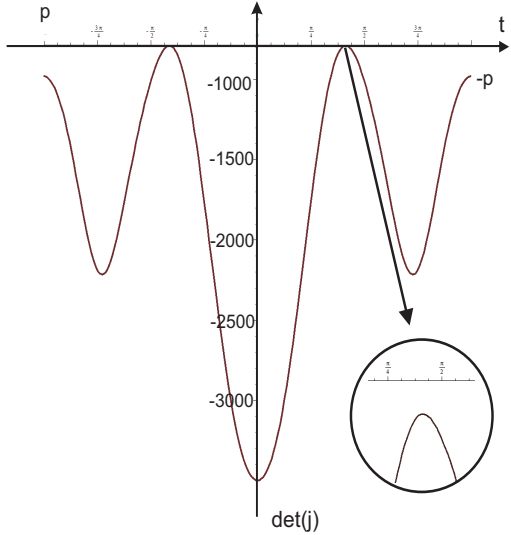} 
       \end{minipage} 
			\end{tabular}
    \end{center}
		\caption{Variation of $\mu_2(\bf t)$ along the Trajectory 2. There are no solutions for $\mu_2(\bf t) = 0 \quad \forall t \in [-\pi, \pi]$.}
		\protect\label{figure:sing12}
\end{figure}

\begin{figure}
    \begin{center}
    \begin{tabular}{@{}c@{}c@{}}
       \begin{minipage}[t]{70 mm}
				\psfrag{t}{$t$}
				\psfrag{det(j)}{$\mu_3(\bf t)$}
				\includegraphics[width=67mm]{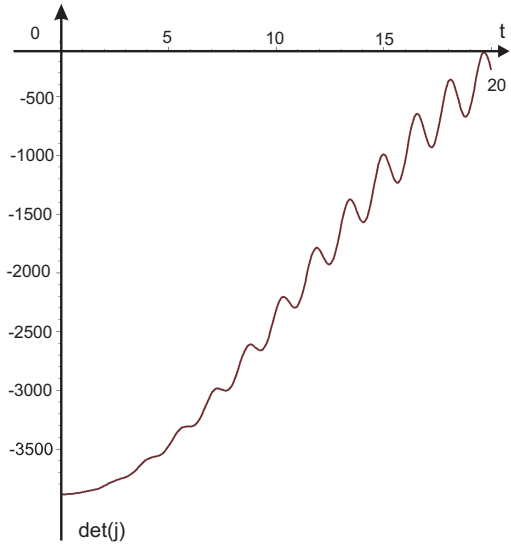} 
       \end{minipage} 
			\end{tabular}
    \end{center}
		\caption{Variation of $\mu_3(\bf t)$ along the Trajectory 3. There are no solutions for $\mu_3(\bf t) = 0 \quad \forall t \in [-\pi, \pi]$.}
		\protect\label{figure:sing13}
\end{figure}

By substituting the values of $\phi_1(t)$, $\phi_2(t)$ and $\phi_3(t)$ from Eq. (\ref{eq:Trajectory1}), Eq. (\ref{eq:Trajectory2}) and Eq. (\ref{eq:Trajectory3}) in $\mu(\bf X)$, we will get $\mu(\phi_1(t))$, $\mu(\phi_2(t))$ and $\mu(\phi_3(t))$ as the determinant of the Jacobian for Trajectory 1, Trajectory 2 and Trajectory 3 respectively (Eq. (\ref{eq:sng})). Due to the large expressions, the equations for  $\mu_1(\bf t)$,  $\mu_2(\bf t)$ and $\mu_3(\bf t)$ are not presented but their roots define the singular configurations : Figures~\ref{figure:sing11}, ~\ref{figure:sing12} and ~\ref{figure:sing13} show the values of these functions when $t$ varies.

\begin{eqnarray}
\label{eq:sng}
  \mu_1(\bf t) = \mu(\phi_1(\bf t)) \quad \quad  \forall t \in [-\pi, \pi]\nonumber \\ 
  \mu_2(\bf t) = \mu(\phi_2(\bf t))  \quad \quad \forall t \in [-\pi, \pi]\nonumber \\
  \mu_3(\bf t) = \mu(\phi_3(\bf t))  \quad \quad \forall t \in [0,~\  20]
  \end{eqnarray}

	The number of solutions of Eq. (\ref{eq:sng}) gives the total number of singular points on the corresponding trajectory. The solutions for $\mu_1(\bf t)$, $\mu_2(\bf t)$ and $\mu_3(\bf t)$ are shown in Eq. (\ref{eq:sol}). From Eq. (\ref{eq:sol}) we get that there exists four solutions for $\mu_1(\bf t)$ and zero solutions for $\mu_2(\bf t)$ and $\mu_3(\bf t)$, which confirms the presence of singular points on Trajectory 1 whereas Trajectory 2\&3 are singularity-free trajectories. Note that numerical approximations are given for readability, but, in practice, isolating intervals with rational bounds with arbitrary width can be computed (see section METHODOLOGY or \cite[chapter 8]{MKCbook09}
 for more details) so that the result is certified.

\begin{eqnarray}
\label{eq:sol}
 \bf t = [-1.51, -0.97, 0.97, 1.51] \gets \mu_1(\bf t) = 0   \quad \forall t \in [-\pi, \pi]\nonumber \\ 
\bf t = [0.97, 1.51] \gets \det({\bf A}) = 0   \quad \forall t \in [-\pi, \pi]\nonumber \\   
 \bf t = [\quad] \gets \mu_2(\bf t) = 0  \quad \forall t \in [-\pi, \pi]\nonumber \\
 \bf t = [\quad] \gets \mu_3(\bf t) = 0   \quad \forall t \in [0,~\ 20]
  \end{eqnarray}

\begin{figure}
    \begin{center}
    \begin{tabular}{@{}c@{}c@{}}
       \begin{minipage}[t]{72 mm}
				\psfrag{z}{$z$}
				\psfrag{x}{$x$}
				\psfrag{y}{$y$}
				\psfrag{tre}{Trajectory 2}
				\psfrag{tra}{Trajectory 1}
				\psfrag{s1}{$s_4$}
				\psfrag{s2}{$s_3$}
				\psfrag{s3}{$s_2$}
				\psfrag{s4}{$s_1$}
				\psfrag{sing}{Singular Points}
				\psfrag{spur}{Spurious}
				\psfrag{sin}{$\xi(\bf X)$}
				\includegraphics[width=68mm]{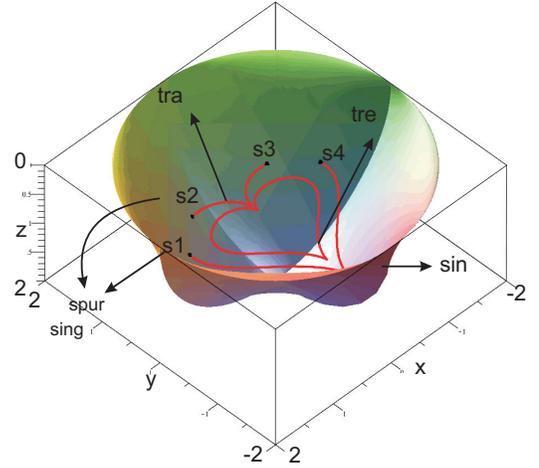} 
       \end{minipage} 
			\end{tabular}
    \caption{Representation of the trajectories with singularity surface. Trajectory 1 cuts the parallel singularity surface $\xi(\bf X)$ in four points $s_1$, $s_2$, $s_3$ and $s_4$ in the workspace of the Orthoglide. Also, it can be seen that Trajectory 2 lies inside $\xi(\bf X)$ as $\mu_2(\bf t) \neq 0 \quad \forall t \in [-\pi, \pi]$.  }
    \protect\label{figure:sing_tra}
    \end{center}
\end{figure}

\begin{figure}
    \begin{center}
    \begin{tabular}{@{}c@{}c@{}}
       \begin{minipage}[t]{72 mm}
				\psfrag{z}{$z$}
				\psfrag{x}{$x$}
				\psfrag{y}{$y$}
			  \psfrag{traj}{Trajectory 1}
				\psfrag{s3}{$s_3$}
				\psfrag{s4}{$s_4$}
				\psfrag{sin}{$\xi(\bf X)$}
				\includegraphics[width=68mm]{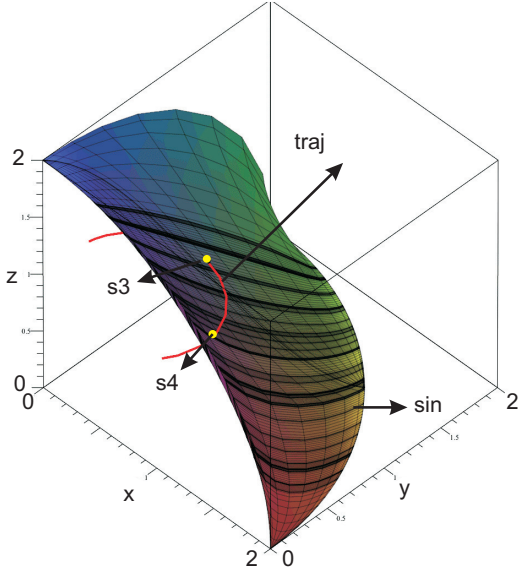} 
       \end{minipage} 
			\end{tabular}
    \caption{Representation of the real singular points $s_3$ and $s_4$ along with the singularity surface $\xi(\bf X)$, which is associated with one working mode.}
    \protect\label{figure:sing1_tra1}
    \end{center}
\end{figure}

$\mu_1(\bf t)$ is the determinant of the Jacobian corresponding to the Trajectory 1. By solving a zero-dimensional system of equations,  it can be shown that  Trajectory 1 cuts the singularity surface in four points $s_1$, $s_2$, $s_3$ and $s_4$ (see Fig.~\ref{figure:sing11} and Eq. (\ref{eq:sol})). The image of these points in workspace is computed by substituting the values of Eq. (\ref{eq:sol}) in Eq. (\ref{eq:Trajectory1}). Similarly, the image of these points in joint space is computed by substituting the values of Eq. (\ref{eq:sol}) in Eq. (\ref{eq:Projection1}). The obtained values are tabulated in Table~\ref{table:sing}. Figures~\ref{figure:sing_tra} represents the Trajectory 1\&2 along with singularity surface $\xi(\bf X)$, also, all the singular points $ \bf s_1$, $\bf s_2$, $\bf s_3$ and $\bf s_4$ are located on the singularity surface.

\begin{figure}
    \begin{center}
    \begin{tabular}{@{}c@{}c@{}}
       \begin{minipage}[t]{70 mm}
				\psfrag{s1}{$s_1$}
				\psfrag{s2}{$s_2$}
				\psfrag{s3}{$s_3$}
				\psfrag{s4}{$s_4$}
				\psfrag{t}{$t$}
				\psfrag{jj}{${\rm det}({\bf A})$}
				\psfrag{p}{$-\pi$}
				\psfrag{pp}{$\pi$}
				\includegraphics[width=67mm]{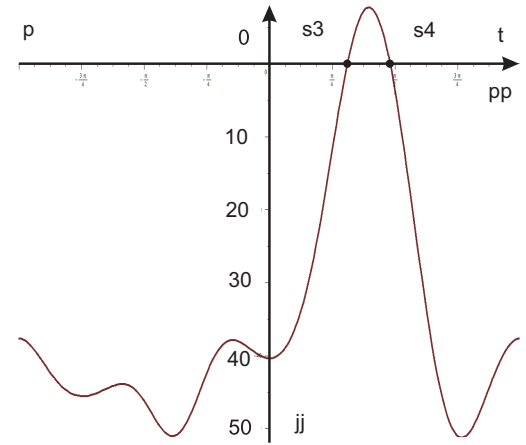} 
       \end{minipage} 
			\end{tabular}
    \end{center}
		\caption{Variation of ${\rm det}({\bf A})$ along the Trajectory 1. $s_3$ and $s_4$ represents the two solutions of ${\rm det}({\bf A}) = 0 \quad \forall t \in [-\pi, \pi]$.}
		\protect\label{figure:sing111}
\end{figure}

$ \bf s_1$ and $\bf s_2$ are the spurious singular points which were introduced by the projection of ${\rm det}({\bf A})$ in Cartesian space. Substituting  $\mu_1(\bf t)$ and $\phi_1(\bf t)$ in ${\rm det}({\bf A})$ gives two solutions ( Eq. (\ref{eq:sol})). These two solutions  $\bf s_3$ and $\bf s_4$ are the real singular points on Trajectory 1. The variation of ${\rm det}({\bf A})$ along Trajectory 1 is shown in Fig.~\ref{figure:sing111}. 

\begin{table}[h!]
\begin{center}
    \caption{Singular postures on Trajectory 1}
    \protect\label{table:sing}
    \begin{tabular}{|c|c|c|c|c|c|c|}
    \hline
		\multicolumn{4}{|c|}{Workspace} & \multicolumn{3}{|c|}{Joint space}\\ \hline
		$S$   & $x$     & $y$   & $z$    & $\rho_1$    & $\rho_2$     & $\rho_3$     \\ \hline
		$S_1$ & $ -1.13$ & $0.35$ & $ 1.00$  & $ 0.55$  &   $1.66$ & $ 2.60$  \\ \hline
		$S_2$ & $ -0.65$ & $0.80$ & $ 1.00$  & $ 0.88$  &   $2.40$ & $ 2.71$  \\ \hline
		$S_3$ & $  0.65$ & $0.80$ & $ 1.00$  & $ 2.18$  & 	$2.40$ & $ 2.71$  \\ \hline
    $S_4$ & $  1.13$ & $0.35$ & $ 1.00$  & $ 2.83$  &   $1.66$ & $ 2.60$  \\ \hline
    \hline
    \end{tabular}
\end{center}
\end{table}
The true singular postures $\bf s_3$ and $\bf s_4$, belonging to the singularity surface $\xi(\bf X)$, which is associated with the working mode used by the Orthoglide prototype ( Fig.~\ref{figure:sing1_tra1}). $\mu_2(\bf t)$ and $\mu_3(\bf t)$ are the determinanst of the Jacobians corresponding to the Trajectories 2\&3 respectively.

The variation of $\mu_2(\bf t)$ along Trajectory 2 for $t \in [-\pi, \pi]$ is shown in Fig.~\ref{figure:sing12}. Figure~\ref{figure:sing13} represents the variation of $\mu_3(\bf t)$ for $t \in [0, 20]$ along Trajectory 3. All the values given in Table~\ref{table:sing} are associated with the trajectories which is marked as $1$ in Fig.~\ref{figure:projection_traj12}. These values are obtained by solving a zero-dimensional system of polynomial equations using RootFinding:-Isolate so that the related values for $sin\_t$ and $cos\_t$ are certified.

In Table~\ref{table:sing}, $\bf s_1$ and $\bf s_2$ are the spurious singular points and $\bf s_3$ and $\bf s_4$ are the real singular points on the Trajectory 1. Trajectory 2\&3 are the singularity-free trajectories.

\section*{Conclusions}
This  paper reports the use of algebraic methods to check the feasibility of given trajectories in the workspace. The general method uses the projection of the polynomial equations associated with the trajectories in the joint space using a Gr\"{o}bner based elimination strategy. There is a significant change in the shape of workspace and the singularity surfaces  due to the joint constraints. The proposed method enables us to project  the joint constraints in the workspace of the manipulator, which further helps to analyse the  projection of the singularities in the workspace with the joint constraints. Such computations ensure the existence of a singular configuration between two poses of the end-effector unlike other numerical or discretization methods. This paper highlights that the singularity analysis should be done in the cross product of the workspace and joint space for parallel robots with several assembly and working modes. The single analysis of the singularities in a projection space can introduce spurious singularities. In fact, the algebraic tools does not allow to distinguish the parallel singularities according to the working mode.
\begin{acknowledgment}
The work presented in this paper was partially funded by the Erasmus Mundus project ``India4EU II''.
\end{acknowledgment}

\bibliographystyle{asmems4}

\end{document}